\documentclass[a4paper]{llncs}
\usepackage[utf8]{inputenc}
\usepackage[english]{babel}
\usepackage{amssymb,amsfonts,amsmath}
\usepackage{graphicx}
\usepackage{alltt}
\usepackage{multicol}
\usepackage{verbatim}
\usepackage{algorithmic}
\usepackage{algorithm}
\usepackage{xcolor}
\usepackage{synttree}
  \branchheight{0.3in}
  \childattachsep{1em}
  \childsidesep{1em}

\newcommand{\url}[1]{{\tt #1}}

\newcommand{\KILL}[1]{}
\newcommand{\HIDE}[1]{}

\newenvironment{datatype}{$\begin{array}{lcl}}{\end{array}$}

\newcommand{\is}{&\ ::=\ &}

\newcommand{\alter}{\ \mid\ }
\newcommand{\altis}{\\ &\ \mid\ &}

\newcommand{\baction}{\\ & & \left\{\begin{array}{lcl}}
\newcommand{\eaction}{\end{array}\right.}
\newcommand{\defby}{& := &}

\newcommand{\nat}{\mathbb{N}}
\newcommand{\binder}[2]{\underset{#2}{#1}}

\begin{document}

\title{First Steps of an Approach to the ARC Challenge based on Descriptive Grid Models and the Minimum Description Length Principle}

\author{Sébastien Ferré}

\institute{Univ Rennes, CNRS, IRISA\\ 
  Campus de Beaulieu, 35042 Rennes, France\\
  Email: \email{ferre@irisa.fr}}

\maketitle

\begin{abstract}
  The Abstraction and Reasoning Corpus (ARC) was recently introduced
  by François Chollet as a tool to measure broad intelligence in both
  humans and machines. It is very challenging, and the best approach
  in a Kaggle competition could only solve 20\% of the tasks, relying
  on brute-force search for chains of hand-crafted transformations.
  In this paper, we present the first steps exploring an approach
  based on descriptive grid models and the Minimum Description Length
  (MDL) principle. The grid models describe the contents of a grid,
  and support both parsing grids and generating grids. The MDL
  principle is used to guide the search for good models, i.e. models
  that compress the grids the most.
  We report on our progress over a year, improving on the general
  approach and the models. Out of the 400 training tasks, our
  performance increased from 5 to 29 solved tasks, only using 30s
  computation time per task. Our approach not only predicts the output
  grids, but also outputs an intelligible model and explanations for
  how the model was incrementally built.
\end{abstract}

\HIDE{
Measure of Intelligence
Artificial Intelligence
Explainable AI
Program Synthesis
Structured Prediction
Minimum Description Length
2D Parsing
}

\sloppy

\section{Introduction}
\label{intro}

This document describes our approach to the ARC (Abstraction and
Reasoning Corpus) challenge introduced by François Chollet as a tool
to measure intelligence, both human and artificial~\cite{Chollet2019}.
ARC was introduced in order to foster research on Artificial General
Intelligence (AGI)~\cite{Goertzel2014agi}, and to move from
system-centric generalization to developer-aware
generalization. The latter enables a system to ``handle situations
that neither the system nor the developer of the system have
encountered before''. For instance, a system trained to recognize cats
features system-centric generalization because it can recognize cats
that it has never seen, but in general, it does not feature
developer-aware generalization because it can not learn to recognize
horses with only a few examples (unlike a young child would).

ARC is made of 1000 learning tasks, 400 for training, 400 for
evaluation, and 200 kept secret by its author in order to ensure
bias-free evaluation. Each task consists in learning how to generate
an output colored grid from an input colored grid (see
Figure~\ref{fig:task} for an example).
ARC is a challenging target for AI. As F. Chollet writes: ``to the
best of our knowledge, ARC does not appear to be approachable by any
existing machine learning technique (including Deep
Learning)''~\cite{Chollet2019}. The main difficulties are the
following:
\begin{itemize}
\item The expected output is not a label, or even a set of labels, but
  a colored grid with size up to 30x30, and with up to 10 different
  colors. It therefore falls in the domain of {\em structured
    prediction}~\cite{dietterich2008structured}.
\item The predicted output has to match exactly the expected
  output. If a single cell is wrong, the task is considered as
  failed. To compensate for that, three attempts are allowed for each
  input grid.
\item In each task, there are generally between 2 and 4 training
  instances (input grid + output grid), and 1 or 2 test instances for
  which a prediction has to be made.
\item Each task relies on a distinct transformation from the input
  grid to the output grid. In particular, no evaluation task can be
  solved by reusing a transformation learned on the training
  tasks. Actually, each task is a distinct learning problem, and what
  ARC evaluates is broad generalization and few-shot learning.
\end{itemize}
A compensation for those difficulties is that the the input and output
data are much simpler than real-life data like photos or texts. We
think that this helps to focus on the core features of intelligence
rather than on scalability issues.

A kaggle
competition\footnote{\url{https://www.kaggle.com/c/abstraction-and-reasoning-challenge}}
was organized in Spring 2020 to challenge the robustness of ARC as a
measure of intelligence, and to see what could be achieved with
state-of-the-art techniques. The winner's solution manages to solve
20\% over 100 (secret) evaluation tasks, which is already a remarkable
performance. However, it is based on the brute-force search for chains
of grid transformations, chosen from a hand-crafted set of 142
elementary transformations. This can hardly scale to larger sets of
tasks, and as the author admits: ``Unfortunately, I don't feel like my
solution itself brings us closer to AGI.'' Another competitor used a
similar domain-specific language based on transformations, and
grammatical evolution to search the program
space~\cite{Fischer2020}. Their approach solves 3\% of the Kaggle
tasks, and 7.68\% over the 400 training tasks.

Our approach is based on the MDL (Minimum Description Length)
principle~\cite{Rissanen1978,Grunwald2019} according to which {\em ``the best
model for some data is the one that compresses the most the data''}.
The MDL principle has already been applied successfully to pattern
mining of various data structures (transactions~\cite{KRIMP2011},
sequences~\cite{sqs2012}, relational databases~\cite{rdb_krimp2009},
or graphs~\cite{BarCelFer2020ida}), as well as
classification~\cite{ProLee2020}.
A major difference with those work is that the model to be learned
needs not only explain the data but also be able to perform a
structured prediction, here the output grid from the input grid.
At this stage (Version~2.2), we consider rather simple models that
have no chance to cover a large proportion of ARC tasks: single-color
points and shapes, and basic arithmetic of positions and sizes. Our
prime objective is to demonstrate the effectiveness of an MDL-based
approach to the ARC challenge, and hopefully to AGI. Our approach
succeeds on 29/400 (7.25\%) training tasks and 6/400 (1.5\%)
evaluation tasks, with a learning timeout of 30s only per task. We
find such a result quite encouraging given the difficulty of the
problem. Moreover, the solved tasks are quite diverse, and the
learned model is often close to the simplest model for the task (in
the chosen class of models).

This report is organized as follows. Section~\ref{arc} gives a formal
definition of ARC grids and tasks. \HIDE{Section~\ref{related} discusses
related work on artificial intelligence, in particular structured
prediction, AGI, or programming by demonstration; it also presents the
MDL principle and some of its applications to AI and knowledge
discovery.} Section~\ref{modelling} explains our modelling of the ARC
tasks in the framework of two-parts MDL, in particular our class of
models. Section~\ref{learning} describes the MDL-based learning
process, which boils down to defining description lengths and model
refinements. Section~\ref{eval} reports on the evaluation of our
approch on training and evaluation ARC tasks, as well as on the
learned models for a few tasks.

\section{Defining ARC Tasks}
\label{arc}

\begin{figure}[t]
  \centering
  \includegraphics[width=\textwidth]{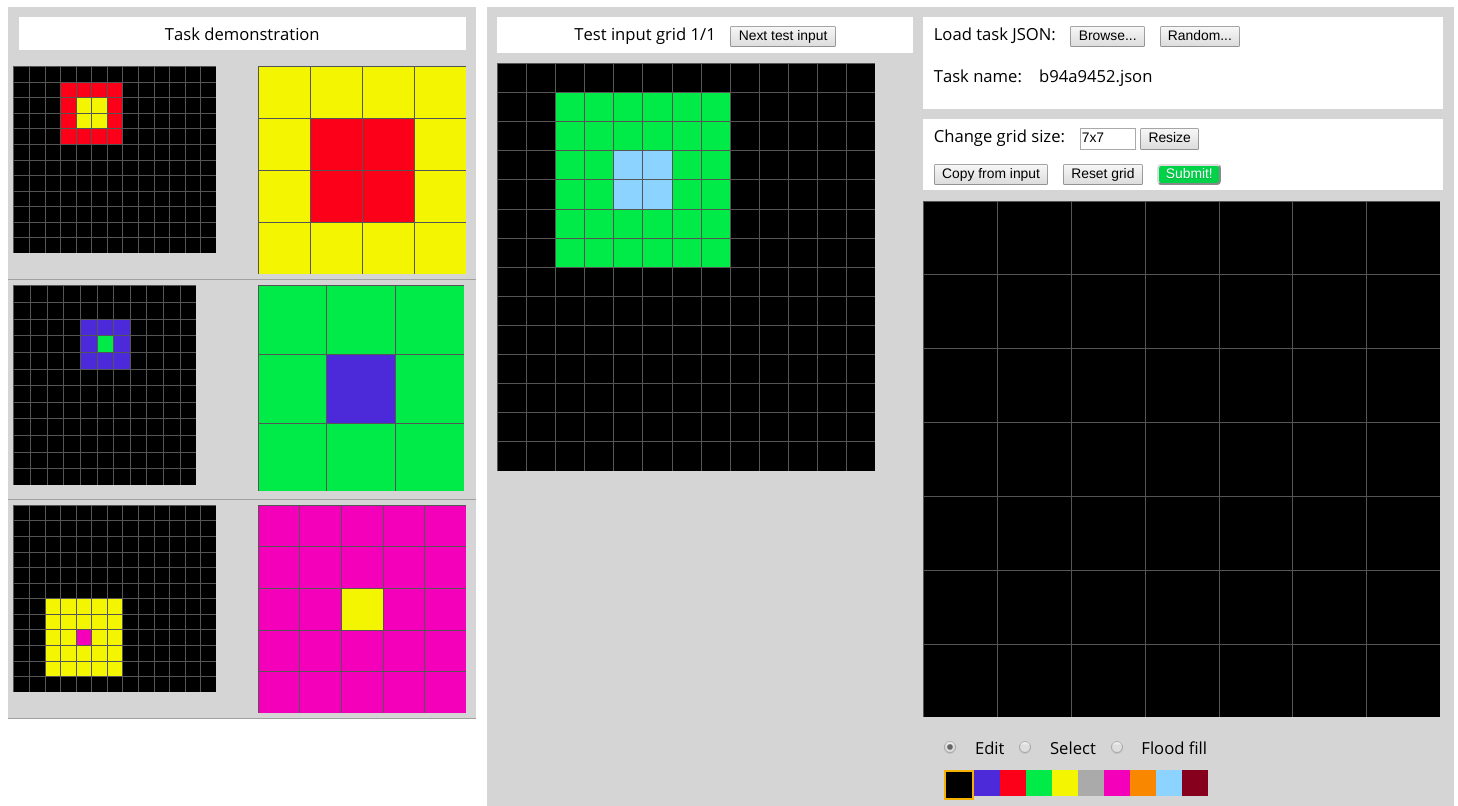}
  \caption{One of the training tasks ({\tt b94a9452}), with three
    examples on the left, a test input grid in the middle, and the
    widget on the right that can be used to draw the missing output
    grid.}
  \label{fig:task}
\end{figure}

\begin{definition}[colors]
  We assume a finite set~$C$ of distinct symbols, which we call {\em colors}.
\end{definition}

In ARC, there are 10 colors coded by digits (0..9) in JSON files, and
by colors (e.g., black, purple, red) in the web interface (see
Figure~\ref{fig:task}).

\begin{definition}[grid]
  A {\em grid} $g \in C^{h \times w}$ is a matrix of colors with $h>0$
  rows, and $w>0$ columns. A grid is often displayed as a colored
  grid. The number of rows~$h = {\it height}(g)$ is called the {\em
    height} of the grid, and the number of
  columns~$w = {\it width}(g)$ is called the {\em width} of the grid.
  
  A {\em grid cell} is caracterized by {\em coordinates}~$(i,j)$,
  where $i$ selects a row, and $j$ selects a column. The color at
  coordinates~$(i,j)$ is denoted either by~$g_{ij}$ or by~$g[i,j]$.
  Coordinates range from~$(0,0)$ to~$(h-1,w-1)$.
\end{definition}

In ARC, grids have a size from $1 \times 1$ to $30 \times 30$.

\begin{definition}[example] 
  An {\em example} is a pair of grids~$e = (g^i,g^o)$, where $g^i$ is called
  the {\em input grid}, and $g^o$ is called the {\em output grid}.
\end{definition}

As illustrated by Figure~\ref{fig:task}, the output grid needs not
have the same size as the input grid, it can be smaller or bigger.

\begin{definition}[task]
  A {\em task} is a pair~$T = (E,F)$, where $E$ is the set of {\em
    train examples}, and $F$ is the set of {\em test examples}.
\end{definition}

The learner has only access to the input and ouput grids of the
train examples. Its objective is, for each test example, to predict
the output grid given the input grid.
As illustrated by Figure~\ref{fig:task}, the different input grids of
a task need not have the same size, nor use the same colors. The same
applies to test grids.

The ARC is composed of 1000 tasks in total: 400 for training, 400 for
evaluation by developers, and 200 for independent
evaluation. Figure~\ref{fig:task} shows one of the 400 training tasks,
with three train examples, and one test input grid. Developers should
only look at the training tasks, not at the evaluation tasks, which
should only be used to evaluate the broad generalization capability of
the system.
Each task has generally between 2 and 4 train examples, and 1 or 2
test examples.

\HIDE{
\section{Related Work}
\label{related}

This work is related to three domains, all part of the field of
Artificial Intelligence: {\em structured prediction} (as a sub-domain
of machine learning), {\em programming by example}, and the {\em
  Minimum Description Length (MDL)} principle.

{\em Structure prediction} is the sub-domain of supervised machine
learning where the outputs to predict are structured objects rather
than scalar values (discrete or
numeric)~\cite{ReviewStructPred,dietterich2008structured}. The output
grids of the ARC challenge are an example. Other common and recent
examples are found in natural language processing (NLP) and generative
media. Examples in NLP are parsing (predicting a parse
tree)~\cite{PredParseTree} or machine translation (predicting a word
sequence)~\cite{MachTransl}. Examples in generative media often use
generative adversarial networks (GAN) to generate photo-realistic
images, music songs or texts~\cite{GAN_App}. However, the previous
examples are not part of supervised machine learning because there is
not one (or a few) correct answer, the generated output ``only'' has
to be realistic to human perception. Structured prediction appears to
be mostly applied to sequences. Indeed, predicting a sequence of
symbols can be decomposed into repeatedly predicting the next
symbol. We could not find work predicting grid structures similar to
ARC grids.

Programming by example, ILP, ...

MDL work...
}

\section{Modelling ARC Tasks in Two-Parts MDL}
\label{modelling}

The objective is here to design a class of {\em task models} that can
be used to represent a transformation from input grids to output
grids.
Our key intuition is that a task model can be decomposed into two grid
models, one for the input grids, and another for the output grids. The
{\em input grid model} expresses what all train input grids have in
common, while the {\em output grid model} expresses what all train
output grids have in common, as a function of the input grid.
The actual transformation from input grid to output grid involves the
two models as follows. The input grid model is used to {\em parse} the
input grid, which produces some information that characterizes the
input grid. Then, that parsing information is fed into the output grid
model in order to generate an output grid.

In the following, paragraphs starting with {\em Version~X} are
specific to that version, in contrast to other paragraphs that are
generic.

\subsection{Models}
\label{model}

\begin{figure}[t]
\begin{center}
\begin{datatype}
  {\it Pair} \is {\bf InOut}(in: {\it Grid}, out: {\it Grid})
  \\
  {\it Grid} \is {\bf Grid}(size: {\it Vector},\ color: C,\ layers: list({\it Object}))
  \\
  {\it Object} \is {\bf PosShape}(pos: {\it Vector},\ shape: {\it Shape})
  \\
  {\it Shape} \is {\bf Point}(color: C)
  \altis {\bf Rectangle}(size: {\it Vector},\ color: C,\ mask: {\it Mask})
  \\
  {\it Vector} \is {\bf Vec}(i: \nat,\ j: \nat)
  \\
  {\it Mask} \is {\bf Bitmap}(bitmap: M)
  \altis {\bf Full} \alter {\bf Border}
  \altis {\bf EvenCheckboard} \alter {\bf OddCheckboard}
  \altis {\bf PlusCross} \alter {\bf TimesCross}
\end{datatype}
\end{center}
\caption{Datatypes used in grid models and grid data}
\label{fig:patterns}
\end{figure}

\paragraph{Version 2.}
The main component of our grid models is {\em objects}. Indeed, ARC
grids can often be read as objects with various shapes over a
background. Objects can be disconnected, nested or overlap each
other. Each object has a position and a shape.

A shape has a color, and either it is a point, or it fits into a
rectangular box of some size. In the latter case, the precise shape is
specified by a mask that can be a custom bitmap or one of a few
pre-defined regular shapes. The position of objects is relative to the
top-left cell of the shape. Each object attribute (e.g., position,
size, color) may be constant (e.g., the shape is always red) or
variable across the examples.

We define our models as templates over data structures representing
concrete pairs of grids. Figure~\ref{fig:patterns} defines those data
structure with algebraic data types for gird pairs, grids, objects,
shapes, masks, and vectors (2D positions and sizes). They use
primitive types for natural numbers ($\nat$), colors ($C$), and 2D
bitmaps (${\it M}$). Uppercase names in bold are called {\em
  constructors}, and the lowercase names of constructor arguments are
called {\em fields}.

A task is made of two grids, one for the input grid, and another for
the output grid.
Each grid is described as a stack of layers on top of a background,
each layer being made of a single object. The background has a size (a
2D vector), and a color.
An object is a shape at some position. A shape is either a point,
described by its color; or a rectangle, described by its size, color,
and mask. That mask allows to account for regular and irregular
shapes, and indicates which pixels in the rectangle box actually
belong to the shape (other pixels can be seen as transparent).
A mask can be a custom bitmap or one of a few common regular shapes:
full rectangle, rectangle border, checkboards, and centered crosses.
A data structure describing the first train pair in Figure~\ref{fig:task}
is:
\[\begin{array}{l}
    {\bf InOut}( \\
    \quad {\bf Grid}({\bf Vec}(12,13), black,\\
    \quad \quad [{\bf PosShape}({\bf Vec}(2,4), {\bf Rectangle}({\bf Vec}(2,2), yellow, {\bf Full})),\\
    \quad \quad ~{\bf PosShape}({\bf Vec}(1,3), {\bf Rectangle}({\bf Vec}(4,4), red, {\bf Full})])), \\
    \quad {\bf Grid}({\bf Vec}(4,4), yellow, \\
    \quad \quad [{\bf PosShape}({\bf Vec}(1,1), {\bf Rectangle}({\bf Vec}(2,2), red, {\bf Full})]))).
  \end{array} \]

A {\em path} in a data structure is a sequence of fields that goes
from the root of the data structure to some sub-structure. In the
above example, path~$in.size.j$ refers to the input grid width, which
is $13$; path~$out.layers[0].shape.color$ refers to the color of the top
shape in the output grid. which is $red$; and path~$in.layers[1].pos$
refers to the position of the bottom shape in the input grid, which is
${\bf Vec}(1,3)$. Given a data structure~$d$ and a path~$p$, the
substructure of~$d$ that is refered by~$p$ is written with the
dot-notation $d.p$.

In order to generalize from specific pairs of grids to general task
models, we introduce two kinds of elements that can be used in place
of any sub-structure: unknowns and expressions.
An {\em unknown}, which we note with a question mark $?$, indicates
that any data structure of the expected type can fit in. For example,
a grid model that matches all input grids in Figure~\ref{fig:task} is:
\[\begin{array}{l}
    {\bf Grid}({\bf Vec}(12,?), black,\\
    \quad [{\bf PosShape}(?, {\bf Rectangle}(?, ?, ?)),\\
    \quad ~{\bf PosShape}(?, {\bf Rectangle}(?, ?, {\bf Full}))])
  \end{array}\]
It matches grids with 12 rows and a black background, and two stacked
rectangles of any position, any size, and any color. The top rectangle
may have any mask, while the bottom rectangle should be full. The
purpose of parsing is to fill in the holes, replacing unknowns with
data structures such that the whole resulting data structure correctly
describes a grid or a pair of grids.
An {\em unknown path} is a path in the model that leads to an
unknown. The unknown paths in the above grid model are: $size$,
$layers[0].pos$, $layers[0].shape.size$, $layers[0].shape.color$,
$layers[0].shape.mask$, $layers[1].pos$, $layers[1].shape.size$,
$layers[1].shape.color$. The set of unknown paths of a model~$m$ is
denoted by~$U(m)$.

A data structure that may contain unknowns is called a {\em
  pattern}. A data structure is said {\em ground} if it contains no
unknown. Given a ground data structure~$d$ and a pattern~$p$, the
notation $d \sim p$ says that $d$ {\em agrees with}~$p$ or that $p$
{\em matches}~$d$. It means that the ground data structure~$d$ can be
obtained by a substitution of the pattern unknowns with ground data
structures. For example, we have
${\bf Vec}(12,14) \sim {\bf Vec}(12,?)$.

An {\em expression} defines part of the data structure as the result
of a computation over some input data structure, which we call the
{\em environment}. Expressions are made of primitive values,
constructors, operators/functions, and variables. In Version~2, the
only operators are zero, addition and substraction over natural numbers,
they are used to compute the position and size of grids and shapes.
Variables are references to parts of the environment. A convenient way
to refer to such parts is to use {\em paths} in the environment data
structure.
Expressions that would only be made of primitive values and
constructors would actually be data structures, and are therefore not
considered as expressions.
The environment of the output grid is the input grid, so that parts of
the output grid can be computed based on the input grid contents. The
input grid has no environment so that expressions are not useful in
the input grid model.
Re-using the above model for input grids, we can now define a correct
and complete model for the task in Figure~\ref{fig:task}, using
unknowns for the input grid, and expressions for the output grid:
\[\begin{array}{l}
    {\bf InOut}( \\
    \quad {\bf Grid}({\bf Vec}(12,?), black,\\
    \quad \quad [~{\bf PosShape}({\bf Vec}(?,?), {\bf Rectangle}({\bf Vec}(?, ?), ?, {\bf Full})),\\
    \quad \quad ~~{\bf PosShape}({\bf Vec}(?,?), {\bf Rectangle}({\bf Vec}(?, ?), ?, {\bf Full})~])), \\
    \quad {\bf Grid}(layers[1].shape.size, layers[0].shape.color, \\
    \quad \quad [~ {\bf PosShape}(\\
    \quad \quad \quad \quad {\bf Vec}(layers[0].pos.i - layers[1].pos.i,\ layers[0].pos.j - layers[1].pos.j), \\
    \quad \quad \quad \quad {\bf Rectangle}(layers[0].size,\ layers[1].color,\ {\bf Full})) ~])).
  \end{array} \]

In words, this models says: ``find two stacked full rectangles on a
black background in the input grid, then generate an output grid,
whose size is the same as the bottom shape, and whose background color
is the color of the top shape, and finally add a full rectangle whose
size is the same as the top shape, whose color is the same as the
bottom shape, and whose position is the difference between the top
shape position and the bottom shape position (relative position).''

A model is {\em well-formed} if it is well-typed (constructor
arguments, operator operand types, etc.), and if all variables are
valid paths in the environment data. More specifically, a task model
is well-formed if the input grid model contains no variable, and if
all variables in the output grid model are valid paths in the input
grid model.

A task model is {\em definite} if its output grid model contains no
unknown, so that it can be used to deterministically generate an
output grid from a data structure describing the input grid.

\subsection{Data According to a Model}
\label{data}

In MDL-based approaches, it is essential to have a lossless encoding
of data in order to have a fair comparison of different
models. Therefore, it is important to identify the ``data according to
the model'', i.e. the data that needs to be added to a model in order
to capture all the information in the original data.

We first consider grid models, where data is a grid. An analogy can be
made with formal languages and grammars. In this analogy, grids are
sentences, grid models are grammars, and data according to the model
is the parse tree. A parse tree explains how the grammar generates the
sentence. Here, data structures as defined in
Figure~\ref{fig:patterns} play the role of parse trees for grid
models, they are {\em grid parse trees} and we denote them with the
greek letter~$\pi$. A grid parse tree, the data according to the
model, is a data structure that instantiates the grid model by
replacing expressions by their value, and unknowns by values such that
the resulting data structure correctly describes the grid. The
description length will only count the data structures that
instantiate the unknowns because the rest of the parse tree is already
known from the model.

For input grid~$g^i_k$ ($k$-th example), $\pi^i_k$ denotes one
possible grid parse tree according to some fixed input grid
model~$m^i$. Similarly, for output grid~$g^o_k$, $\pi^o_k$ denotes one
possible grid parse tree according to some fixed output grid
model~$m^o$ and output environment~$\varepsilon^o$.

In practice, a grid model may not explain all the cells of a
grid. This happens in case of noise in the data, and also in the
learning process where intermediate models are obviously
incomplete. To account for those situations without losing
information, we define ``data according to the model'' as a grid parse
tree plus the list of cells that differ between the original grid~$g$
and the grid generated by the grid parse tree~$\pi$. We call this
additional information a {\em grid delta}, written~$\delta$.
\KILL{Formally, a grid
delta is a set of triples~$(i,j,c) \in \nat \times \nat \times C$
stating that the cell at coordinates~$(i,j)$ has color~$c$ in~$g$,
which is different from the color specified by~$\pi$.}

An additional difficulty is when a test input grid does not match the
input grid model learned from the train examples. For instance, all
input grids in the train examples have size (10,10) but the test
input grid has size (10,12). In this case, there is no grid parse tree
that agrees with both the grid model and the grid. In many cases, this
discrepancy does not alter the validity of the model, which only
happens to be overspecific (a form of overfitting).
Our approach to this difficulty is to allow for approximate parsing,
i.e. to allow grid parse trees to diverge from the grid model. As
those differences are present in the grid parse tree~$\pi$, there is
no loss of information. Of course, the description length will take
into acount such differences to penalize them.

As an illustration, let us consider the following (incomplete) input
grid model for the task in Figure~\ref{fig:task}:
\[ {\bf Grid}(?, black, [\ {\bf PosShape}(?, {\bf Rectangle}(?,?,{\bf Full}))\ ]). \] There
are mainly two ``data according to the model'' for the first input
grid~$g^i_1$ (the upper index is $i$ for input grids and $o$ for
output grids, the lower index identifies the example):
\begin{enumerate}
\item matching the outer red rectangle\\
  $\pi^i_1 = {\bf Grid}({\bf Vec}(12,13), black,$\\
  \hspace*{2cm} $[\ {\bf PosShape}({\bf Vec}(1,3), {\bf Rectangle}({\bf Vec}(4,4), red, {\bf Full}))\ ])$\\
  $\delta^i_1 = \{ (2,4,yellow), (2,5,yellow), (3,4,yellow), (3,5,yellow) \}$

\item matching the inner yellow rectangle\\
  $\pi^i_1 = {\bf Grid}({\bf Vec}(12,13), black,$\\
  \hspace*{2cm} $[\ {\bf PosShape}({\bf Vec}(2,4), {\bf Rectangle}({\bf Vec}(2,2), yellow, {\bf Full}))\ ])$\\
  $\delta^i_1 = \{ (1,3,red), (1,4,red), (1,5,red), (1,6,red), (2,3,red), (2,6,red), (3,3,red), ...\}$
\end{enumerate}
The first solution looks better because it has a much smaller grid
delta, and its grid parse tree has the same size as in the second
solution.

\subsection{Primitive Functions on Grids, Grid Models, Grid Parse Trees, and Grid Deltas}
\label{primitive:functions}

We here define the key primitive functions to manipulate grids, grid
models, grid parse trees and grid deltas.

We first define {\em grid deltas} and simple arithmetics on them.

\begin{definition}[grid delta]
  Let $g_1, g_2 \in C^{h \times w}$ two same-size grids. The {\em grid
    delta} $\delta \subseteq \nat \times \nat \times C$ between the
  two grids, written $g_2 - g_1$, is defined as the set of cells that
  need changing color in~$g_1$ in order to obtain~$g_2$.
  \[ \delta = g_2 - g_1 = \{ (i,j,c) \mid g_2[i,j] \neq g_1[i,j], c = g_2[i,j] \} \]
  
  A grid delta~$\delta$ can be added to a grid~$g_1$ to get a corrected same-size grid~$g_2$.
  \[ g_2 = g_1 + \delta \iff \forall i \in [0,h[, j \in [0,w[: g_2[i,j] =
    \left\{
      \begin{array}{ll}
        c & \quad\textrm{if~} (i,j,c) \in \delta \\
        g_1[i,j] & \quad\textrm{otherwise} \\
      \end{array}\right. \]
\end{definition}

Then we define two functions that are directly related to the
semantics of grid models. The first function converts a grid parse tree
into a grid.

\begin{definition}[grid drawing]
  Let $\pi$ be a grid parse tree, ${\it draw}(\pi)$ is the grid that
  results from the drawing following the instructions given by the
  grid parse tree.  
\end{definition}

\paragraph{Version 2.} Given a grid parse tree~$\pi$ as specified in
Figure~\ref{fig:patterns}, the grid ${\it draw}(\pi)$ is obtained by
first generating a grid with the specified size and background color,
and then layers are drawn on that grid, bottom-up. Each layer, either
a point or a rectangle, is drawn in the obvious way. For a point
${\bf PosShape}({\bf Vec}(i,j), {\bf Point}(c))$, the cell at
position~$(i,j)$ is set to color~$c$. For a rectangle
${\bf PosShape}({\bf Vec}(i,j), {\bf Rectangle}({\bf Vec}(h,w),c,m))$,
each cell at position $(i+x,j+y)$, for any $x \in [0,h[$ and
$y \in [0,w[$, is set to color~$c$ if the relative position~$(x,y)$
belongs to mask~$m$.

\paragraph{}
The second function applies a grid model to an
environment, resolving references to it, and replacing expressions by
their evaluation result.

\begin{definition}[model application]
  Let $m$ be a grid model, and $\varepsilon$ an environment for the
  model. ${\it apply}(m,\varepsilon)$ is the model that results from
  the substitution of expressions in~$m$ by their evaluation over the
  environment~$\varepsilon$. The resulting model has no expression but
  may still have unknowns.
  The operation is ill-defined if some variable in~$m$ is not defined
  in~$\varepsilon$. 
\end{definition}

\paragraph{Version 2.} Environments are grid parse trees, and
variables are paths into them. There are only three operators -- zero, plus,
minus -- and they are evaluated in the usual way on integer values.

\paragraph{}
We also define two non-deterministic functions that produce grid parse
trees from a grid model, and hence grid descriptions. The first
function corresponds to the parsing of a grid.

\begin{definition}[grid parsing]
  Let $m$ be a grid model, and $g$ be a grid. ${\it parse}(m,g)$
  non-deterministically returns a grid parse tree~$\pi$ that
  approximately agrees with model~$m$, and approximately draws as
  grid~$g$. Non-determinism allows for different interpretations of
  the grid by the model.
\end{definition}

\paragraph{Version 2.} This is by far the most complex primitive
function.
Parsing is decomposed in three stages. First, the grid is partitioned
into contiguous monocolor regions.
Second, objects are looked for as single parts or as unions of parts
to cover the case where some objects are overlaped by other
objects. Small parts (less than 5 covered cells) are also considered
as adjacent individual points. When an object's shape is not a point
or a full rectangle, two shapes are considered: one shape with a full
mask, considering the missing cells as noise on top of the rectangle;
and another shape with a mask containing the cells in the rectangle
box that belong to the object and have the same color. A regular mask
is used when possible (e.g., borders, crosses), otherwise a bitmap is
used.
Third, matching superpositions of objects are looked for according to
the grid model. For each layer, candidate objects are considered in a
fixed ordering that includes some simple heuristics. Black objects are
considered last because black is often the background color. Then,
objects are considered in decreasing number of covered cells. Beyond
those heuristics, a total ordering is defined because it makes
experiments more reproducible, and also because it helps disambiguate
some situations. For instance, if the model contains several points,
they will be matched to points in the grid from left to right,
top-down.
Layers are parsed top-down. When done in a naive recursive way, the
search space is not be explored in fair way. Indeed, the second
candidate object for the frst layer is only considered when all
candidates have been considered for the other layers in combination
with the first candidate object for the first layer. Defining the rank
of a parse tree for all layers as the sum of the rank of the chosen
candidate object of each layer, we explore the search space in
increasing rank to favor the first candidate objects across all
layers. This improvement was introduced in Version~2.2.

As an example of parsing, the test input grid in Figure~\ref{fig:task}
has three contiguous parts: the light blue one, the green one, and the
black one. Three rectangles can be identified from those parts: a
small light blue square, a green rectangle overlaped by the blue one,
and a black rectangle covering the whole grid and overlaped by the
other rectangles. According to the above model, this input grid is
parsed as a 2x2 light blue rectangle at coordinates (3,4), on top of a
6x6 green rectangle at coordinates (1,2), on top of a black
background.

Compared to Version~1, parsing has been made non-deterministic in the
sense that a sequence of grid parse trees is produced. We use the
description lengths as defined below to order this sequence from the
shorter DL to the longer DL. Indeed, the shorter the DL, the more
plausible the grid parse tree is. Because of combinatorial problems in
the parsing process, we use various bounds: maximum number of
candidate objects for a given layer in the model, maximum number of
produced grid parse trees before sorting them, and maximum number of
grid parse trees actually returned. Relative to approximate parsing,
we also bound the number of differences w.r.t. the model.

\paragraph{}
The second function corresponds to generation, where the grid parse
tree is produced by filling in the holes of the grid model,
i.e. guessing at the missing elements.

\begin{definition}[grid generation]
  Let $m$ be a grid model without any expression. ${\it generate}(m)$
  non-deterministically returns a grid parse tree that is obtained by
  replacing unknowns in the model by type-compatible sub-trees.
\end{definition}

\paragraph{Version 2.} We simply replace unknowns by default
values. It makes it very unlikely to generate the right output grid by
chance but it helps to visualize the intermediate states of the
learning process. The default position is $(0,0)$, the default size is
$(10,10)$ for grids, and $(2,2)$ for rectangles, the default color is
black for grids and grey for shapes, the default mask is the full
mask, the default shape is a rectangle with default values in each
field.

\subsection{Reading and Writing a Grid with a Grid Model}
\label{read:write}

We define two important components for reading and writing grids.
They are simply composed from the above primitve functions, and are
used as basic blocks for training and using models.

\begin{figure}[t]
  \centering
  \begin{minipage}[b]{0.55\textwidth}
    \centering
    \includegraphics[height=3.2cm]{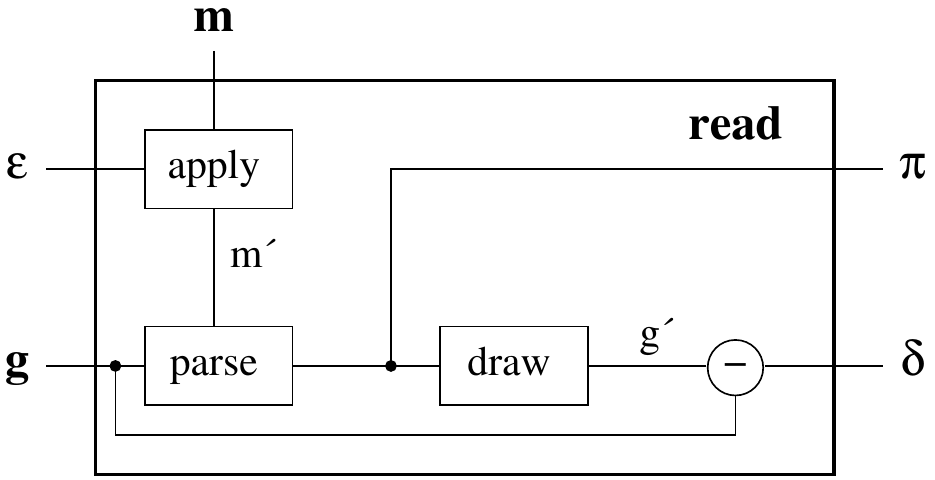}
    \caption{The {\em read} component.}
    \label{fig:read}
  \end{minipage}
  \hfill
  \begin{minipage}[b]{0.44\textwidth}
    \centering
    \includegraphics[height=3.2cm]{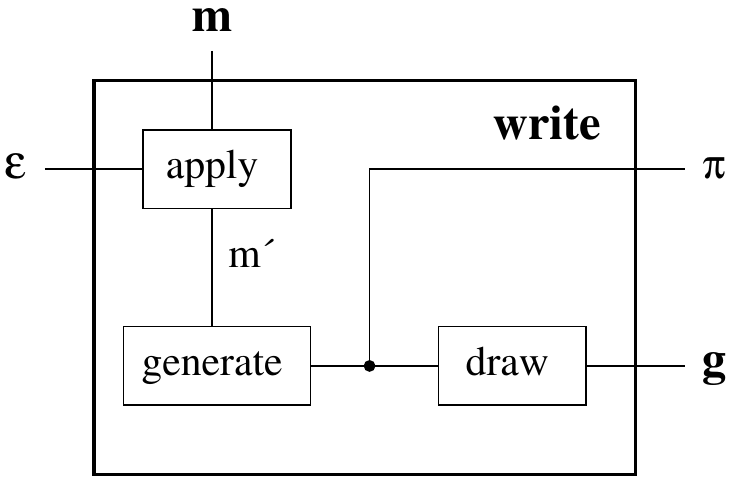}
    \caption{The {\em write} component.}
    \label{fig:write}
  \end{minipage}
\end{figure}

\begin{definition}[grid reading]
  Let $m$ be a model. Function ${\it read}_m$ takes an environment and
  a grid as input, and returns a grid parse tree and delta as output.
  \[ {\it read}_m(\varepsilon,g) = \{ (\pi,\delta) \mid m_\varepsilon
    = {\it apply}(m,\varepsilon), \pi \in {\it
      parse}(m_\varepsilon,g), \delta = g - {\it draw}(\pi) \} \]
\end{definition}

Figure~\ref{fig:read} gives a schematic description of the {\em read}
component. When reading a grid~$g$, the grid model is first applied to
the input environment, in order to resolve any expression that the
model may contain. The resulting model is then used to parse the input
grid into a grid parse tree~$\pi$. Finally, the grid delta between the
grid specified by~$\pi$ and~$g$ is computed. The pair~$(\pi,\delta)$
provides a lossless representation of the input grid, as expressed by
the following lemma.
\[ (\pi,\delta) \in {\it read}_m(\varepsilon,g) \Longrightarrow g =
  {\it draw}(\pi) + \delta. \] 

\begin{definition}[grid writing]
  Let $m$ be a grid model. Function ${\it write}_m(\varepsilon)$ takes
  an environment as input, and returns a parse tree and a grid as
  output.
  \[ {\it write}_m(\varepsilon) = \{ (\pi,g) \mid m_\varepsilon = {\it
      apply}(m,\varepsilon), \pi = {\it generate}(m_\varepsilon), g =
    {\it draw}(\pi) \} \]
\end{definition}

Figure~\ref{fig:write} gives a schematic description of the {\em
  write} component. When writing a grid, the grid model is first
applied to the input environment, in order to resolve any expression
that the model may contain. The resulting model is then used to
generate a grid parse tree, by instantiating the unknowns, from which
a concrete grid can be drawn. The parse tree is output in addition to
the grid in order to provide an explanation of how the grid was
generated.

\subsection{Task Models: Prediction, Training, and Creation}
\label{training:predicting}


\begin{figure}[t]
  \centering
  \includegraphics[width=0.6\textwidth]{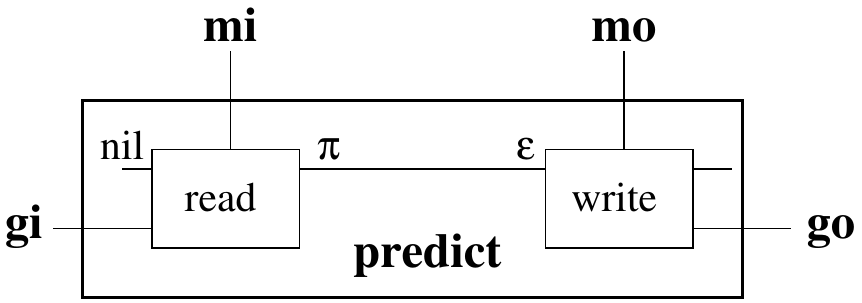}
  \caption{Prediction mode: producing an output grid from the input grid}
  \label{fig:predict}
\end{figure}

Using a task model~$M = {\bf InOut}(m^i,m^o)$ to predict the output
grid from the input grid simply consists into the chaining of grid reading and grid writing.


\[ {\it predict}_M(g^i) = \{g^o \mid (\pi^i,\delta^i) \in {\it read}_{m^i}({\it nil},g^i), g^o \in {\it write}_{m^o}(\pi^i) \} \]

Figure~\ref{fig:predict} gives a schematic description of
prediction. The input grid~$g^i$ is first read with the input grid
model~$m^i$, using an empty environment, which non-deterministically
results in an input grid parse tree~$\pi^i$ and an input
delta~$\delta^i$. Then, the predicted output grid is produced by
writing the output grid model~$m^o$ using the input grid parse tree as
environment.

For instance, given the input grid parse tree returned by the reading
of the test input grid in Figure~\ref{fig:task}, the output grid model
generates a 2x2 green rectangle at coordinates (2,2), on top of a 6x6
light blue background, which is the expected grid.

\begin{figure}[t]
  \centering
  \includegraphics[width=0.6\textwidth]{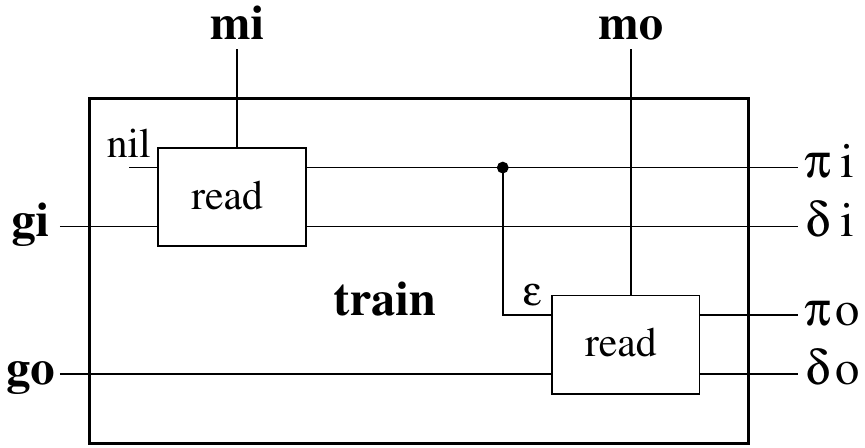}
  \caption{Training mode: reading the input and output grids in chain}
  \label{fig:train}
\end{figure}

\paragraph{}
During the training phase, both the input and output grids are
known. However, computing a grid delta between the predicted output
grid and the expected output grid is not enough to learn a better
model. It is more useuful to have the grid parse tree for each grid,
because it provides an explanation of how the model ``sees'' the
grids. We therefore define the ${\it train}$ function that chains the
reading of the input and ouput grids.

\HIDE{ On output grids~$y_k$, we use parsing rather than
  generating. This is explained by the fact that description lengths
  are used at the learning stage rather than at the application
  stage. During learning, output models are incomplete, and still
  require some parameters and delta to generate the output
  grid~$y_k$. However, this output grid is available during learning
  (for train examples), and using parsing enables to find which
  parameters and delta are required. The objective of learning is to
  make those output parameters and delta become empty. Knowing what
  are the required parameters and delta can help the learner improve
  the model in this direction.}

\[\begin{array}{rl}
    {\it train}_M(g^i,g^o) = \{ (\pi^i,\delta^i,\pi^o,\delta^o) \mid & (\pi^i,\delta^i) \in {\it read}_{m^i}({\it nil},g^i),\\
                                                                     & (\pi^o,\delta^o) \in {\it read}_{m^o}(\pi^i,g^o) \}
  \end{array}\]

Figure~\ref{fig:train} gives a schematic description of training. The
first step is the same as prediction. The input~$g^i$ is first read
with the input grid model~$m^i$, using an empty environment, which
non-deterministically resuts in an input grid parse tree~$\pi^i$ and
an input grid delta~$\delta^i$. In the second step, the output
grid~$g^o$ is read with the output grid model~$m^o$, using $\pi^i$ as
environment, which non-deterministically results in an output grid
parse tree~$\pi^o$ and an output grid delta~$\delta^o$.
The two grid parse trees and the two grid deltas collectively form a
lossless description of the two grids according to the task model.

\begin{figure}[t]
  \centering
  \includegraphics[width=0.6\textwidth]{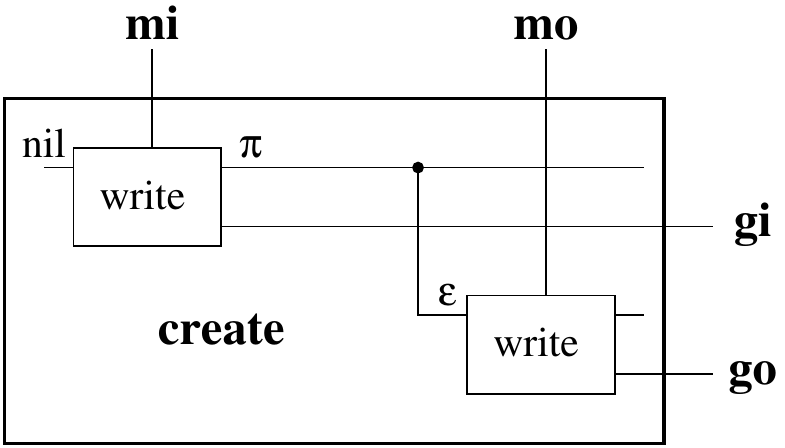}
  \caption{Creation mode: writing input and output grids in chain}
  \label{fig:create}
\end{figure}

\paragraph{}
It is also possible to {\em create} new examples for the task, i.e. to
create both the input and output grids from nothing else than the
input and output grid models.

\[{\it create}_M() = \{ (g^i,g^o) \mid (\pi^i,g^i) \in {\it write}_{m^i}({\it nil}), (\pi^o,g^o) \in {\it write}_{m^o}(\pi^i) \} \]

Figure~\ref{fig:create} gives a schematic description of the creation
mode. The input grid model is used to write the input grid given an
empty environment. The parse tree of the input grid is then used as an
environment for the output grid model to generate an output grid that
is consistent with the input grid. The process is highly
non-deterministic as in general many grids can be generated for a
given input grid model. Indeed, it generally contains several unknowns.

\section{Learning Task Models}
\label{learning}

MDL-based learning works by selecting the model that compresses the
data the best. This involves two main components: (a) a measure of
compression so as to compare models, and (b) a strategy to explore the
model space. The measure of compression is called {\em description
  length}.

\subsection{Description Lengths}
\label{dl}

In two-part MDL, the description length (DL)~$L(M,D)$ measures the number
of bits required to communicate data~$D$ with model~$M$. It is
decomposed in two parts according to the equation
\[ L(M,D) = L(M) + L(D|M) \] where $L(M)$ is the DL of the model, and
$L(D|M)$ is the DL of the data encoded with the model.
In our case, a model~$M$ is a task model~${\bf InOut}(m^i,m^o)$, and
data~$D$ is the part of a task~$T = (E,F)$ that is accessible to
the learner. As the objective is to learn a model that generalizes to
new instances, only the train examples are accessible, so
$D = E$.

The description length of the task model is the sum of the description
lengths of its two grid models.
\[ L(M) = L(m^i) + L(m^o) \]
We detail the description length of grid models below. The
description length of the task data~$D = E$ is the sum of the
description length of each example.
\[ L(D|M) = L(E|M) = \alpha \sum_{(g^i,g^o) \in E} L(g^i|M) + L(g^o|M) \]
Factor~$\alpha$ equals~10 by default to give more weights to the data,
relative to the model, and hence allow the learning of more complex
models. Indeed, ARC tasks have a very low number of examples (3 on
average).

The description length of each grid, according to a task model, is
based on the chained reading of the pair of grids. First, the
description length of a grid relative to a given reading boils down to
description lengths of the grid parse tree and the grid delta.
\[ L(g|m,\varepsilon,\pi,\delta) = L(\pi|{\it apply}(m,\varepsilon)) +
  L(\delta|\pi) \]
Note that the description lengths of the grid parse
tree~$\pi$ is relative to the grid model~$m$ applied to its
environment. Indeed, we only need to encode the parts of the grid
parse tree that are unknown in the grid model or that differ from the
grid model. Its detailed definition depends on the definition of grid
models.
The description length of a grid delta~$\delta$ is relative to a grid
parse tree~$\pi$. This is valid as the grid parse tree is encoded
before the delta, and is therefore available when encoding the grid
delta. Its detailed definition also slightly depends on grid models.

Then, we define the absolute description length of input grids and
output grids by choosing the ``best reading'', i.e. the chained
reading that minimizes the cumulated DL of both grids.
\begin{align*}
  L(g^i|M) & = L(g^i|m^i,{\it nil},\pi^i_*,\delta^i_*) \\
  L(g^o|M) & = L(g^o|m^o,\pi^i_*,\pi^o_*,\delta^o_*) \\
  & \textrm{where~} (\pi^i_*,\delta^i_*,\pi^o_*,\delta^o_*) =\\
  & \quad \binder{\operatorname{argmin}}{(\pi^i,\delta^i,\pi^o,
  \delta^o) \in {\it train}_M(g^i,g^o)} L(g^i|m^i,{\it nil},\pi^i,\delta^i) + L(g^o|m^o,\pi^i,\pi^o,\delta^o)
%
\end{align*}

We also introduce normalized definitions of description
lengths~$\hat{L}$. Indeed, input and output grids may have very
different sizes (e.g., the output grid has a single cell), and this
can hinder the learning of a grid model for the smaller grid. Our
normalization gives the same weight to the input grids and to the
output grids, taking the initial model~$M_0$ as a point of reference.
We therefore define $\lambda^i$ and $\lambda^o$ as respectively the
input and output contributions to $L(M_0,D) = \lambda^i + \lambda^o$.
\begin{align*}
    \lambda^i = L(m^i_0) + \alpha \sum_{(g^i,g^o) \in E} L(g^i|M_0) \\
    \lambda^o = L(m^o_0) + \alpha \sum_{(g^i,g^o) \in E} L(g^o|M_0)
\end{align*}
From there, we define normalized description lengths as follows.
\begin{align*}
  \hat{L}(M,D) = \hat{L}(M) + \hat{L}(D|M) \\
  \hat{L}(M) = \frac{1}{\lambda^i}L(m^i) + \frac{1}{\lambda^o}L(m^o) \\
  \hat{L}(D|M) = \alpha \sum_{(g^i,g^o) \in E} \frac{1}{\lambda^i}L(g^i|M) + \frac{1}{\lambda^o}L(g^o|M) \\
\end{align*}
As the learning strategy only selects models giving shorter and shorter description lengths, the normalized description~$\hat{L}(M,D)$ is in the interval~$[0,2]$, and the contribution of each part (input and output) is in the interval~$[0,1]$.

When defining description lengths for grid models and grid parse
trees, we assume a few basic description lengths for elementary types:
\begin{itemize}
\item $L_\nat(n) = 2\,\log (n+1) + 1$: universal encoding of natural
  intergers (including zero), more precisely the gamma encoding
  shifted by one to include zero;
\item $L_P(x \in X) = -\log P(x)$: encoding based on a probability distribution~$P$ over elements~$X$;
\item $L_X(x \in X) = \log |X|$: encoding based on a uniform distribution over elements~$X$.
\end{itemize}

\paragraph{Version~2.}
In order to define the description lengths of grid models and grid
parse trees, we first need to define the description lengths of
primitive types, algebraic datatypes, expressions, and templates.

{\em Primitive types.} Integers are used for positions and sizes. For
positions, we use a uniform distribution over the rows or columns of a
grid. Hence for a row position~$i$ in a grid~$g$,
$L(i|g) = \log {\it height}(g)$; similarly, for a column position~$j$
in a grid~$g$, $L(j|g) = \log {\it width}(g)$. When the grid size is
not known, the maximum grid size is used (30 in the ARC challenge).
For sizes, we use the universal encoding of natural numbers~$L_\nat$.
For the color of shapes, we use the uniform distribution~$L_C$. For
the background color of grids, we use a probability
distribution~$P_{BC}$ that gives a higher probability to color black
($P_{BC} = 0.91$) than to other colors ($P_{BC} = 0.01$). Indeed, in
most tasks, grids have a black background. Hence, for a background
color~$c$, $L(c) = -\log P_{BC}(c)$. \KILL{We then define the
  description length of background colors $L_{BC}(c)$ as
  $L_{P_{bc}}(c)$.}
The description length of a rectangular bitmap is simply defined as
its number of cells: $L(m) = height(m) \times width(m)$.

{\em Algebraic types.} For each algebraic datatype (${\it Grid}$,
${\it Object}$, ${\it Shape}$, ${\it Vector}$, ${\it Mask}$), each
constructor determines which fields are to be encoded. Therefore, it
suffices to encode the choice of a constructor among the constructors
of the datatype, in addition to the encoding of the fields. For
datatypes~${\it Grid}$, ${\it Object}$, and ${\it Vector}$, there is a
single constructor, hence there is no choice to be made. Hence,
$L({\bf Grid}) = L({\bf Object}) = L({\bf Vec}) = 0$. For
datatype~${\it Shape}$, there are two constructors, to which we give a
uniform distribution. Hence,
$L({\bf Point}) = L({\bf Rectangle}) = -\log 0.5 = 1$.  For
datatype~${\it Shape}$, there are 7 constructors, to which we give the
following distribution:
$P_{\it Shape} = \{ {\bf Bitmap}: 0.3,\ {\bf Full}: 0.5,\ {\bf Border}: 0.1,\ {\bf
  EvenCheckboard}: 0.025,\ {\bf OddCheckboard}: 0.025,\ {\bf PlusCross}:
0.025,\ {\bf TimesCross}: 0.025 \}$.
For a given constructor, the encoding of some field may depend on
other fields, thus inducing an ordering of fields. This is the case
for constructor~${\bf Rectangle}$ where the mask field depends on the
size field (see the encoding of masks above).\footnote{We could also consider
that the position field depends on the size field. Indeed, given a
grid size and a rectangle size, not all positions are possible if we
assume that the rectangle must fit into the grid.}
%
Grids have a list of shapes. The encoding of a list starts with the
universal encoding of the length of the list, followed by the encoding
of the list elements.

{\em Expressions.} There are two kinds of expressions (with
distribution~$P_e$): function (or operator) applications ($P_e = 0.5$,
1 bit), and variables ($P_e = 0.5$, 1 bit). Functions are encoded
along a uniform distribution over the functions whose result type is
the expected type. The expected type is determined by the syntactic
context of the expression: e.g., a color if it is the color field of
the grid or a shape, an integer if it is a component of the position
of a shape, or an operand of an arithmetic operator. Currently, there
are only three functions -- zero, addition, and substraction -- all
with result type ${\it int}$. Variables are encoded according to the
environment signature, i.e. the set of variables that are available in
the environment, grouped by data type. A variable is encoded according
to a probability distribution over the environment variables of same
type. In order to distinguish between the same-type variables, we
define the probability distribution as the softmax of the similarities
between the paths of those variables with the expression path. This
implies that defining a shape height as a function of a shape height
is prefered to defining it as a function of a shape width or a shape
position.

{\em Templates.} There are three kinds of templates (with
distribution~$P_t$): primitive values and constructors ($P_t = 0.4$,
1.3 bits), expressions ($P_t = 0.5$, 1 bit), and unknowns
($P_t = 0.1$, 3.3 bits). Each field of a constructor and each argument
of a function is a template, thus allowing in principle arbitrary
mixing of values, constructors, functions, and unknowns. In Version~2,
we never use unknowns for grids, objects, shapes, and function
arguments. Expressions are given higher probability because they
explain some part of the grid parse tree from the
environment. unknowns are given the lower probability because they
correspond to unexplained parts.

{\em Grid models} ($L(m)$) are templates over type ${\it Grid}$,
and are encoded as such.

{\em Grid parse trees} ($L(\pi|m)$) are only made of primitive values
and constructors, and could be encoded as such. However, as grid parse
trees are obtained by parsing grids according to a grid model, parts
of the grid parse tree are already encoded in the grid
model. Actually, the very purpose of grid models is precisely to
factor out what all example grids have in common, and avoiding to
encode it repeatedly. The only parts of grid parse trees that need to
be encoded are (a) the parts corresponding to unknowns in the grid
model, and (b) the parts that differ from the grid model due to
approximate parsing.
\begin{itemize}
\item[(a)] About the unknowns, a fixed ordering can be derived from
  the model. Therefore, it is enough to concatenate the encodings of
  the grid parse subtrees that correspond to each unknown.
\item[(b)] About the differing parts, we have no a priori information,
  neither on their number, nor on their location in the grid parse
  tree. We first encode their number with universal encoding. We then
  encode each differing part by encoding its location in the grid
  parse tree, using a uniform distribution, and the subtree at that
  location.
\end{itemize}
The encoding of the location of the differing parts could probably be
improved in several ways. First, some differences are more likely than
others: e.g., a different rectangle width vs a completely different
shape. Second, not all subsets of locations make sense as a difference
location should not be inside another difference location as it would
entail to encode the same part twice.

{\em Grid delta} ($L(\delta|\pi)$). For comparability with grid models and grid parse
trees, we encode a grid delta~$\delta$ as a set of points.  The
description length of a grid delta is hence defined as:
\[ L(\delta|\pi) = L_\nat(|\delta|) + \sum_{(i,j,c) \in \delta} L({\bf PosShape}({\bf Vec}(i,j), {\bf Point}(c))|{\it draw}(\pi)). \]
Note that the description length of points is made relative to the
grid drawn from the grid parse tree. In particular, this makes the
grid size available to the encoding of the point position.


\begin{table}[t]
  \centering
  \caption{Comparison of description lengths of the initial model (top) and the solution model given above (bottom).}
    \begin{center}
      \begin{tabular}{|l||r|r||r|r|}
    \hline
          & $L(M)$ & $L(D|M)$ & $L(M,D)$ & $\hat{L}(M,D)$ \\
    \hline
    \hline
    input & 9.0 & 10645.0 & 10653.9 & 1.000 \\
    \hline
    output & 9.0 & 1714.0 & 1723.0 & 1.000 \\
    \hline
    \hline
    chained & 17.9 & 12359.0 & 12376.9 & 2.000 \\
    \hline
      \end{tabular}
    \end{center}
  \begin{center}
    \begin{tabular}{|l||r|r||r|r|}
    \hline
      & $L(M)$ & $L(D|M)$ & $L(M,D)$ & $\hat{L}(M,D)$ \\
    \hline
    \hline
    input & 73.5 & 1711.3 & 1784.9 & 0.168 \\
    \hline
    output & 53.9 & 0.0 & 53.9 & 0.031 \\
    \hline
    \hline
    chained & 127.4 & 1711.3 & 1838.7 & 0.199 \\
    \hline
    \end{tabular}
  \end{center}
\label{tab:dl:example}
\end{table}

\paragraph{}
For the task in Figure~\ref{fig:task}, Table~\ref{tab:dl:example}
compares the description lengths of the initial model and the solution
model given as example above, for $\alpha=10$ (each example counts as
10). The tables detail the split between input grid and output grid on
one hand, and the split between the model ($L(M)$) and the data
according to the model ($L(D|M)$) on the other hand.
We observe that the increase in the model DL is largely compensated by
the decrease of the data DL. The resulting compression gain is 16.3\%, a six-fold
reduction. In particular, with the solution model, the description
length of the ouput grids has fallen to zero, which means that output
grids are entirely explained by the model and the input grid parse
tree.

\KILL{
\paragraph{Version 1.1.} The description length for shapes is extended
in the obvious way, giving the same weight to points and rectangles.

\begin{datatype}
  {\it shape} \is {\bf Point}(i: {\it attr}_\nat, j: {\it attr}_\nat, color: {\it attr}_C)
  \baction
  L(shape,N_V) \defby 1 + L(i,N_V,L_\nat) + L(j,N_V,L_\nat) + L(color,N_V,L_C) \\
  \eaction
  \altis {\bf Rectangle}(height: {\it attr}_\nat, width: {\it attr}_\nat, mini: {\it attr}_\nat, minj: {\it attr}_\nat,\\
  & & \hspace*{1cm} color: {\it attr}_C, mask: {\it attr}_M)
  \baction
  L(shape,N_V) \defby 1 + L(height,N_V,L_\nat) + L(width,N_V,L_\nat) \\
  & + & L(mini,N_V,L_\nat) + L(minj,N_V,L_\nat) + L(color,N_V,L_C) + L(mask,N_V,L_M)
  \eaction
\end{datatype}

\begin{center}
  \begin{tabular}{|l|r|r|r|}
    \hline
    model & $L(m)$ & $L(E|m)$ & $L(m,E)$ \\
    \hline
    initial & 8.0 & 23315.5 & 23323.5 \\
    example & 114.1 & 1788.4 & 1902.6 \\
    \hline
  \end{tabular}
\end{center}
}

\subsection{Learning Strategy}

The model space is generally too vast for a systematic exploration,
and heuristics have to be employed to guide the search for a good
model.
The learning strategy consists in the iterative refinement of models,
starting with an initial model, in search for the best model, i.e. the
most compressive model. Given a class of models, a search strategy is
therefore entirely specified by:
\begin{enumerate}
\item an initial model~$M_0$ ({\em initial model}), generally the
  simplest one;
\item and a function~$R$ mapping each model~$M$ to an ordered
  list~$M_1, M_2, \ldots$ of refined models ({\em refinements}),
  generally taking into account the data according to the model (here,
  grids, grid parse trees and grid deltas).
\end{enumerate}
We say that a refinement~$M_i$ is {\em compressive} if it reduces the
description length compared to the original model~$M$. The ordering of
refinements is important because the number of possible refinements
could be huge, and it has a cost to evaluate each refinement by
measuring its description length. Indeed, this involves the parsing of
both input and output grids by each refined model.  This ordering is
therefore a key part of the heuristics.

Another part of the heuristics consists in bounding the number of
compressive refinements per model that are considered, and the number
of models that are kept after each iteration. A {\em greedy search}
selects the first compressive refinement, and has therefore a single
model at each iteration. A {\em beam search} provides a wider
exploration by selecting $K_r$ compressive refinements per model, and
then selecting among them the $K_m$ best models for the next
iteration. $K_m$ is the beam width. The search halts when no
compressive refinement can be found.

\paragraph{Version 2.} The initial model is made of two minimal grid
models, reduced to a background with unknown size and color, and no
shape layer.
\[ {\bf InOut}({\bf Grid}(?,?,[\ ]), {\bf Grid}(?,?,[\ ])) \]

Two kinds of refinements are used: (1) the addition of a new object to
a grid model (input or output), and (2) the replacement of a part of a
template (typically some unknown) by another template (typically a
value or an expression).
In the first kind of refinement, the new object can be inserted at any
position in the list of layers. The inserted object can be a fully
unspecified point or rectangle: ${\bf PosShape}(?, {\bf Point}(?))$ or
${\bf PosShape}(?, {\bf Rectangle}(?,?,?))$. In the output grid model,
the inserted object can also be an object or a shape from the input
grid model, using paths as variables to reference them: $p_{object}$
or ${\bf PosShape}(?, p_{shape})$. Indeed, it is common for the output
grid to reuse shapes from the input grid, either at the same position
or at a different position to be determined. Inserting such references
to the input grid is not necessary because they could be learned from
unspecified objects but they accelerate learning.

The second kind of refinements enables to replace part of a model at
path~$p$ by a template~$t$. The replacing template is either a {\em
  pattern}, i.e. a combination of primitive values, constructors, and
unknowns; or an {\em expression} using environment variables. For the
input grid model, it can only be patterns because the environment is
empty. For the output grid model, every path into the input grid parse
tree is available as an environment variable.
In this version, we only consider patterns made of a single primitive
value or a single constructor with all fields being unknowns; and we
only consider expressions on integers with a form among $x$, $x+c$,
$x-c$, $x+y$, $x-y$, for any environment variables~$x,y$, and any
constant integer~$c \in \{1,2,3\}$.
When the replaced part~$p$ is an unknown, it enables to make the model
more specific to the task. When the replaced part is a pattern in the
output grid model, and it is replaced by an expression, it enables to
better define the output grid as a function of the input grid. To
summarize, unknowns may be replaced by patterns and expressions, and
patterns in the output grid model may be replaced by expressions.

Let $E$ be the set of examples (pairs of grids), and $M = (m^i,m^o)$
be the current model. We generate the following refinements:
\begin{enumerate}
\item An input unknown at~$p \in U(m^i)$ can be replaced by
  pattern~$t$ if
\[ \forall (g^i_k,g^o_k) \in E: \exists (\pi^i_k,\delta^i_k,\pi^o_k,\delta^o_k) \in {\it
    train}_M(g^i_k,g^o_k): \pi^i_k.p \sim t \]
\item An output unknown at~$p \in U(m^o)$ can be replaced by
pattern~$t$ if
\[ \forall (g^i_k,g^o_k) \in E: \exists (\pi^i_k,\delta^i_k,\pi^o_k,\delta^o_k) \in {\it
    train}_M(g^i_k,g^o_k): \pi^o_k.p \sim t \]
\item An output grid model part at~$p \in m^o$ can be replaced by expression~$e$ if
\[ \forall (g^i_k,g^o_k) \in E: \exists (\pi^i_k,\delta^i_k,\pi^o_k,\delta^o_k) \in {\it
    train}_M(g^i_k,g^o_k): \pi^o_k.p = {\it apply}(e, \pi^i_k) \]
\end{enumerate}

A tentative ordering of refinements is the following: output shapes,
input shapes, output expressions, input expressions.  Among
expressions, the ordering is: $x$, $x-c$, $x+c$, $x-y$, $x+y$.

The sequence of compressive refinements that leads to the above model
about task in Figure~\ref{fig:task} is shown in the table below. The
normalized description length is the one reached after applying the
refinement. Each refinement is described as an equality between a
model path (left-hand side) and a template (right-hand
side). Refinements defining a layer by a shape actually insert a new
layer for that shape.

\begin{center}
  \begin{tabular}{|r|r|l|}
    \hline
    step & $\hat{L}(M,D)$ & refinement \\
    \hline
    0 & 2.000 & \\
    1 & 1.272 & $in.layer[1] = {\bf PosShape}(?, {\bf Rectangle}(?,?,?))$ \\
    2 & 1.031 & $out.size = layer[1].shape.size$ \\
    3 & 0.814 & $out.layer[0] = {\bf PosShape}(?, {\bf Rectangle}(?,?,?))$ \\
    4 & 0.732 & $out.layer[0].shape.color = in.layer[1].shape.color$ \\
    5 & 0.688 & $in.layer[0] = {\bf PosShape}(?, {\bf Rectangle}(?,?,?))$ \\
    6 & 0.549 & $out.color = in.layer[0].shape.color$ \\
    7 & 0.423 & $out.layer[0].shape.size = in.layer[0].shape.size$ \\
    8 & 0.381 & $out.layer[0].shape.mask = in.layer[0].shape.mask$ \\
    9 & 0.361 & $out.layer[0].shape.pos = {\bf Vec}(?,?)$ \\
    10 & 0.310 & $out.layer[0].shape.pos.i = in.layer[0].shape.pos.i - in.layer[1].shape.pos.i$ \\
    11 & 0.259 & $out.layer[0].shape.pos.j = in.layer[0].shape.pos.j - in.layer[1].shape.pos.j$ \\
    12 & 0.252 & $in.layer[0].shape.mask = {\bf Full}$ \\
    13 & 0.246 & $in.layer[1].shape.mask = {\bf Full}$ \\
    14 & 0.241 & $in.color = black$ \\
    15 & 0.238 & $in.size = {\bf Vec}(?,?)$ \\
    16 & 0.212 & $in.size.i = 12$ \\
    17 & 0.209 & $in.layer[0].shape.pos = {\bf Vec}(?,?)$ \\
    18 & 0.205 & $in.layer[0].shape.size = {\bf Vec}(?,?)$ \\
    19 & 0.202 & $in.layer[1].shape.pos = {\bf Vec}(?,?)$ \\
    20 & 0.199 & $in.layer[1].shape.size = {\bf Vec}(?,?)$ \\
    \hline
  \end{tabular}
\end{center}

At steps 1 and 5, two rectangle objects are introduced in the input.
At step 3, a rectangle object is introduced in the output. Combined
with the background grid, this is enough to cover all cells in both
grids. As of step 5, there are therefore empty grid deltas.
At steps 2 and 6, the size and color of the output grid are
respectively equated to the bottom input shape
($in.layer[1].shape.size$) and to the top input shape
($in.layer[0].shape.color$). At steps 4 and 7, the size and color of
the output shape are equated to respectively to the top input shape
($in.layer[0].shape.size$), and to the bottom input shape
($in.layer[1].shape.color$). At step 8, the mask of the ouput shape is
equated to the mask of the top input shape (always ${\bf Full}$ in the
examples). At this stage, only the position of the output shape remains
to be specified.
Steps 9-11 expand the output shape position as a vector, and find the
expressions that define those positions, here differences between the
top input shape and the bottom input shape.
At this stage, the model has solved the task as it can correctly
generate the output grid for any input grid.
Steps 12-13 finds that all input shapes are {\em full}
rectangles. Step 14 finds that the input background color is black.
The other steps expand the remaining positions and sizes as vectors,
and find that all input grids have 12 rows. The latter is an
accidental regularity, and approximate parsing is necessary to parse
the test instance as it has 14 rows.

\section{Evaluation}
\label{eval}

\begin{table}[t]
  \centering
  \caption{Performance on training and evaluation tasks as the number of tasks solved, per version}
  \begin{tabular}{|c|c|c|c|c|c|c|c|c|c|}
    \hline
     & & & & \multicolumn{3}{c|}{training} & \multicolumn{3}{c|}{evaluation} \\
    version & date & $\alpha$ & timeout & time & train & test & time & train & test \\
    \hline
    v1.0 & 2020-10-21 & ~10~ & 20s & 7s & ~8 / 9.5~ & ~5 / 5.5~ & 12s & ~0 / 0.3~ & ~0 / 0.0~ \\
    \hline
    v1.1 & 2021-03-01 & ~10~ & 2s & 1.5s & ~16 / 17.3~ & ~5 / 5.0~ & 1.5s & ~4 / 4.3~ & ~4 / 4.0~ \\
    \hline
    v2.0 & 2021-05-01 & ~10~ & 30s & 20s & ~28 / 34.6~ & ~19 / 19.5~ & 24s & ~6 / 7.8~  & ~5 / 5.0~ \\
    v2.0 & 2021-05-01 & ~10~ & 60s & 35s & ~33 / 41.2~ & ~22 / 22.5~ & 46s & ~7 / 9.9~ & ~6 / 6.0~ \\
    v2.0 & 2021-05-01 & ~10~ & 120s & 57s & ~35 / 43.5~ & ~24 / 24.5~ & 82s & ~8 / 10.9~ & ~7 / 7.0~ \\
    \hline
    v2.1 & 2021-09-12 & ~10~ & 30s & 18s & ~32 / 40.0~ & ~25 / 25.5~ & 23s & ~7 / 10.9~ & ~7 / 7.0~ \\
    v2.1 & 2021-09-12 & ~10~ & 60s & 32s & ~36 / 44.9~ & ~27 / 27.5~ & 43s & ~8 / 12.5~ & ~7 / 7.0~ \\
    v2.1 & 2021-09-12 & ~10~ & 120s & 52s & ~38 / 46.9~ & ~28 / 28.5~ & 77s & ~10 / 14.0~ & ~8 / 8.5~ \\
    \hline
    v2.2 & 2021-10-22 & ~10~ & 30s & 14s & ~41 / 50.1~ & ~29 / 29.5~ & 20s & ~8 / 12.7 & ~6 / 6.5~ \\
    \hline
  \end{tabular}
  \label{tab:eval}
\end{table}

Table~\ref{tab:eval} reports on the performance of our approach on
both training and evaluation tasks (400 tasks each), for successive
versions and timeouts. The timeout sets a limit to the learning time
for each task. On both collections of tasks, we provide performance
for train examples, for which output grids are known, and for test
examples. In each case, we give two measures $n_1/n_2$: $n_1$ is the
number of tasks for which {\em all} examples are correctly solved, and
$n_2$ is the sum of partial scores over tasks. For instance, if a task
has two test examples and only one is solved, the score is~$0.5$.

\paragraph{Version 1.} This first version was designed by looking at
tasks {\tt ba97ae07} and {\tt b94a9452}. Evaluating it on training
tasks has shown that this simple model manages to solve 3.5 additional
tasks: {\tt 6f8cd79b}, {\tt 25ff71a9} (2 correct examples), {\tt
  5582e5ca}, {\tt bda2d7a6} (1 correct example out of 2). This version
even manages to explain all train examples of 3 additional tasks,
hence 8 tasks in total, although they fail on the related test
examples: tasks {\tt a79310a0} (3 examples), {\tt bb43febb} (2
examples), {\tt 694f12f3} (2 examples), and {\tt a61f2674} (2
examples). For task {\tt a79310a0}, v1 recognizes that a shape is
moved, and changes color but, whereas all train shapes are rectangles,
the test shape is not a rectangle. For tasks {\tt bb43febb} and {\tt
  694f12f3}, v1 finds unplausible expressions to replace some
unknowns, e.g. replacing some height by a position instead of a height
decremented by a constant, or replacing some unknown by a constant
that happens to be valid for train examples but not for the test
example. For task {\tt a61f2674}, v1 is not powerful enough but given
that there are only two train examples, it manages to find some ad-hoc
regularity.

Despite those encouraging result on training tasks, no evaluation task
could be solved with this version. We wonder whether this is a
statistical fluke or the fact that evaluation tasks are harder than
the training tasks.

\begin{table}[t]
  \centering
  \caption{Performance on training tasks as the number of tasks solved, for different strategies, in Version~1.1 with timeout 2s, and $\alpha=10$. C = constants first, V = variables first, Ei = input expressions, Si = input shapes, Eo = output expressions, So = output shapes.}
  \begin{tabular}{|c|c|c|c|}
    \hline
    strategy & train & test \\
    \hline
    V/Si-Ei-So-Eo & ~11 / 11.7~ & ~4 / 4.0~ \\
    V/Ei-Si-So-Eo & ~14 / 15.0~ & ~5 / 5.0~ \\
    V/Ei-Si-Eo-So & ~16 / 17.3~ & ~5 / 5.0~ \\
    C/Ei-Si-Eo-So & ~16 / 17.3~ & ~5 / 5.0~ \\
    \hline
  \end{tabular}
  \label{tab:v11:strategies}
\end{table}

\paragraph{Version 1.1.}  Table~\ref{tab:v11:strategies} compare
results with Version~1.1 for different strategies concerning model
refinements. For example, the first line consider expressions with
variables before constants, and consider refinements in this order:
input shapes, input expressions, output shapes, output expressions.
The best strategy among those tested is Ei-Si-Eo-So. Refining the
input model first provides useful parameters when refining the output
model. Inserting expressions before inserting new shapes helps to
disambiguate the parsing process, which often happens when a model
contains several unspecific shapes.

The successful training tasks are {\tt ba97ae07}, {\tt bda2d7a6}, {\tt
  6f8cd79b}, {\tt e48d4e1a}, {\tt 25ff71a9}. Compared to Version~1, we
improve on {\tt bda2d7a6} by succeeding on the the two test instances,
and gain {\tt e48d4e1a}. The former is explained by the use of masks,
and the latter is explained by the improved extraction of rectangles
from grids.
However, we loose on {\tt b94a9452} and {\tt
  5582e5ca}. In Version~1, the latter was successful by chance, and
the former because test inputs were considered for training.
Finally, 17.3 training instances are successful compared to 9.6 in the
previous version. Hence, all in all, we consider this new version as a
valuable improvement.

This is confirmed by Table~\ref{tab:eval} that shows that this new
version succeeds in 4 evaluation tasks, instead of none for the
previous version.

\paragraph{Version 2.} Table~\ref{tab:eval} reports results for
Version~2 for different timeouts. It uses a greedy search ($K_m = 1$)
but at each step, it look for up to 20 compressive refinements
($K_r = 20$), selecting the best one among them. Grid parsing looks
for up to 512 grid parse trees, and keep only the three best
ones. Approximate parsing is not activated on train examples, and
limited to 3 differences on test examples.
Table~\ref{tab:eval} shows a sharp improvement w.r.t. previous
versions, especially on the training tasks. The number of successful
tasks jumps from~5 to 24 in the training dataset, and from~4 to~8 in
the evaluation dataset. This seems to confirm that evaluation tasks
are intrisically more complex than training tasks.
As the language of models is the same as in Version~1.1, the
improvement come mostly from the way those models are used and
modified:
\begin{itemize}
\item the use of paths in grid parse trees rather than variable names,
\item the introduction of non-determinism in grid parsing,
\item the ordering of candidate grid parse trees and refinements,
\item the ability to replace any part of a template, not only unknowns,
\item better definitions of description lengths.
\end{itemize}
Higher timeouts are required because of the non-determinism that
significantly increases computation complexity.

For a timeout of 60s, the 22 successful training tasks are: {\tt
  1bfc4729}, {\tt 1cf80156}, {\tt 1f85a75f}, {\tt 25ff71a9}, {\tt
  445eab21}, {\tt 48d8fb45}, {\tt 5521c0d9}, {\tt 5582e5ca}, {\tt
  681b3aeb}, {\tt 6f8cd79b}, {\tt a1570a43}, {\tt a79310a0}, {\tt
  a87f7484}, {\tt aabf363d}, {\tt b1948b0a}, {\tt b94a9452}, {\tt
  ba97ae07}, {\tt bda2d7a6}, {\tt bdad9b1f}, {\tt e48d4e1a}, {\tt
  e9afcf9a}, {\tt ea32f347}.  For a timeout of 120s, the additional
successful tasks are: {\tt 23581191}, {\tt 91714a58}.
Task~{\tt 48d8fb45} is successful although the learned model does not
exhibit a proper understanding of it. The system found a correlation
that was probably not intended by the creator of the task.
The models learned on other tasks are adequate, and are rather diverse
despite the simplicity of the model language. We informally describe a
few learned models to illustrate this diversity.
\begin{itemize}
\item {\tt a79310a0}. Find a cyan rectangle and move it one row
  down. Unexpectedly, the test instance has an irregular shape instead
  of a rectangle. Thanks to approximate parsing, the input grid parse
  tree contains a mask of the irregular shape, which is used to
  generate the output grid.
\item {\tt aabf363d}. Find an arbitrary shape and a colored point in
  the bottom left corner, and output the same shape except for the
  color, which is the same as the point.
\item {\tt b94a9452}. Find an arbitray shape on a pink background,
  simply change the background color to red.
\item {\tt ba97ae07}. Find two full rectangles, reverse the list of
  layers: bottom rectangle on top, top rectangle at bottom.
\item {\tt bdad9b1f}. Find two segments, a 2x1 cyan rectangle, and 1x2
  red rectangle. Extend each segment into a line spanning the whole
  grid, then add a yellow point at the crossing of the two lines.
\item {\tt e48d4e1a}. Find two colored lines, a 1x10 rectangle and a
  10x1 rectangle, on a 10x10 black grid. Find also a grey rectangle in
  the top right corner with width=1. Generate the same lines at
  different positions, according to the height~$H$ of the grey
  rectangle. The vertical line is moved $H$ columns left
  (substraction), and the horizontal line is moved $H$ rows down
  (addition). This involves arithmetic expressions for computing the
  new positions.
\item {\tt ea32f347}. Find three grey full rectangles, implictly
  ordered from the larger (top layer) to the smaller (bottom layer)
  according to the parsing ordering heuristics. Generate the same
  rectangles with different colors: the top one gets blue, the middle
  one gets yellow, and the bottom one gets red.
\item {\tt 1cf80156}. Crop an arbitrary colored shape on a black grid.
\item {\tt 1f85a75f}. Crop an arbitrary colored shape on a black grid
  that also contains many points of other colors at random positions
  (a kind of noise).
\item {\tt 6f8cd79b}. Starting with a black grid of any size, generate
  a cyan grid of same size, and add a black rectangle at position
  (1,1) whose size is two rows and two columns less than the grid. The
  effect is to add a cyan border to the input grid.
\item {\tt 445eab21}. Find two shapes, the larger one implicitly on
  top. Generate a 2x2 grid whose color is the same as the top shape.
\item {\tt 681b3aeb}. Find two shapes on a black background. Generate
  a 3x3 grid with the color of one shape, and add at position (0,0)
  the other shape. Actually, the two shapes complement each other and
  pave the 3x3 grid. The learned model works because the evaluation
  protocol allows for three predicted output grids, and there are only
  two possibilities as the top left cell must belong to one of the two
  shapes. Moreover, as the larger shape is tried first and it has a
  higher probability to occupy the top left cell, there is more than
  50\% chance being correct on first trial.
\item {\tt 5521c0d9}. Find three full rectangles: a yellow one, a red
  one, and a blue one. Move each of them upward in the grid by as much
  as their height.
\item {\tt 5582e5ca}. Find two shapes on a 3x3 colored grid. Generate
  the same colored grid without the shapes. Actually, the task is to
  identify the majority color in the input grid. Even without a notion
  of counting and majority in the models, the MDL principle implicitly
  identifies the majority color as the one that compresses the most.
\end{itemize}
\HIDE{
  \item {\tt 05269061}. Find three shapes, and generate a three-color
  checker boards as a combination of three shapes with complex
  masks. One color is handled by the background color. Another color
  is handled by the top shape whose mask is a set of stripes. The last
  color is handled by the two bottom shapes, although a single one
  should be sufficient. The three input shapes are three stripes
  picked at random, one for each color. The learned model cannot
  predict for sure which color is where, so that it often fails on
  first trial. This task can actually not be genuinely understood by
  Version~2 models but interestingly, it could nonetheless be solved
  to some extent.}

For 11 training tasks, the approach was successful on all train
instances although it failed on the test instances. This means that a
working model was found but that it does not generalize to the test
instances. Here are some causes for the lack of generalization:
\begin{itemize}
\item wrong choice between constants and variables, and finding
  accidental arithmetic equalities (e.g., {\tt 0962bcdd});
\item wrong segmentation of shapes, e.g. prefering a full rectangle to
  an irregular shape (e.g., {\tt 1caeab9d});
\item the transformation should be performed several times in the test
  instance whereas once is enough in all train instances (e.g., {\tt
    29c11459});
\item there is an invariance by grid rotation or grid transposition
  (e.g., {\tt 4522001f}).
\end{itemize}

\paragraph{Versions 2.1, 2.2.} This version brings only a few changes to the
model but it significantly improves the efficiency of parsing grids,
which is the major cost of learning a model. Table~\ref{tab:eval}
shows that we achieve similar results in 30s as we did in 120s. The
limit of the number of parse trees per grid was lowered from 512 down
to 64 thanks to a better parsing strategy for stacks of layers.

For the record, the main changes relative to Version~2 are the following:
\begin{itemize}
\item distinction between shapes and objects to better represent object moves,
\item richer set of masks (borders, checkboards, and crosses),
\item zero as a nullary operator to avoid expressions like $e_1 - e_1$,
\item small parts can be seen as adjacent points,
\item better strategy when parsing a stack of layers,
\item normalized description length to balance inputs and outputs w.r.t. grid size,
\item insertion of input shapes and objects into the output model,
\item optimizations to reduce timeout.
\end{itemize}

The 7 newly solved training tasks w.r.t. Version~2 at 60s are: {\tt
  08ed6ac7}, {\tt 23581191}, {\tt 72ca375d}*, {\tt 7e0986d6}, {\tt
  a61ba2ce}, {\tt be94b721}, {\tt ddf7fa4f}*. In the two tasks with a
star, the learned model is not precise enough but because the system
is allowed three attempts and there are only a few reading
alternatives, it finds the correct answer (without really
understanding why). For example, in task~{\tt 72ca375d}, the input
grid has three objects, and the output grid has only one of them, the
one that has a symmetry. Our system only understands that the output
selects an input object, and makes an attempt for each object.

\section{Discussion and Perspectives}
\label{conclu}


We have demonstrated that combining descriptive grid models and the
MDL principle is a viable and encouraging approach to the ARC
challenge. In contrast to other approaches, it focuses on the contents
of grids rather than on transformations from input grids to output
grids. Actually, our learning algorithm does not even try to predict
the output grid from the input grid, only to find a short joint
description of the pair of grids, only assuming that the output grid
may depend on the input grid. Prediction of the output grid only comes
as a by-product. The learned model can also be used to create new
examples. We think it makes our approach more robust and scalable, and
more cognitively plausible.

The progress through the different versions has more to do with
improvements in the general approach than with the class of
models. Indeed, the model has not changed much from the first version
(stacks of points and boxed shapes over a colored background, plus
very simple arithmetics). The key to improvements is {\em
  flexibility}. Flexibility lies mostly in the parsing procedure
through non-determinism and approximation. To tackle the complexity
coming from those, the MDL principle appears as essentiel to select
the most plausible parsings, i.e. those that have the shortest
description length. The MDL principle is therefore used at two levels:
at a low level for parsing grids, and at a high level for choosing
model refinements.

We were surprised to see that results on the evaluation tasks are much
lower than on the training tasks. The same observation has been made
previously~\cite{Fischer2020}. We refrained to look at the evaluation
tasks, as advised by the ARC designer, to avoid the inclusion of
ARC-specific priors in our approach. We suspect that the evaluation
tasks have larger or more complex grids because the runtime per task
is higher. We also suspect that they require more difficult
generalizations as this is a key factor of the measure of intelligence
proposed by F. Chollet.

We are confident that our approach can solve many more tasks because
the current grid models are quite simple. First, only a few of the
core human priors are covered. Second, a number of tasks involve
relatively simple whole-grid transformations (e.g. rotations,
symmetries). Those are typically the tasks that are solved by
transformation-based approaches but are not yet solvable by our
model. Another important limit is when a grid contains a
variable-sized collection of objects, each of which must be treated in
the same way. This requires some form of loop in the model.


\begin{thebibliography}{10}
\providecommand{\url}[1]{\texttt{#1}}
\providecommand{\urlprefix}{URL }

\bibitem{BarCelFer2020ida}
Bariatti, F., Cellier, P., Ferr{\'{e}}, S.: Graphmdl: Graph pattern selection
  based on minimum description length. In: Berthold, M.R., Feelders, A.,
  Krempl, G. (eds.) Advances in Intelligent Data Analysis - 18th International
  Symposium on Intelligent Data Analysis, {IDA}. LNCS, vol. 12080, pp. 54--66.
  Springer (2020), \url{https://doi.org/10.1007/978-3-030-44584-3\_5}

\bibitem{Chollet2019}
Chollet, F.: On the measure of intelligence. arXiv preprint arXiv:1911.01547
  (2019)

\bibitem{dietterich2008structured}
Dietterich, T.G., Domingos, P., Getoor, L., Muggleton, S., Tadepalli, P.:
  Structured machine learning: the next ten years. Machine Learning  73(1),
  3--23 (2008)

\bibitem{Fischer2020}
Fischer, R., Jakobs, M., M\"{u}cke, S., Morik, K.: Solving {Abstract}
  {Reasoning} {Tasks} with {Grammatical} {Evolution}. In: {LWDA}. pp. 6--10.
  CEUR-WS 2738 (2020)

\bibitem{Goertzel2014agi}
Goertzel, B.: Artificial general intelligence: concept, state of the art, and
  future prospects. Journal of Artificial General Intelligence  5(1), ~1 (2014)

\bibitem{Grunwald2019}
Gr{\"u}nwald, P., Roos, T.: Minimum description length revisited. arXiv
  preprint arXiv:1908.08484  (2019)

\bibitem{rdb_krimp2009}
Koopman, A., Siebes, A.: Characteristic {Relational} {Patterns}. In: {ACM}
  {SIGKDD} {Int.} {Conf.} {Knowledge} {Discovery} and {Data} {Mining}. pp.
  437--446. {KDD} '09, ACM (2009)

\bibitem{ProLee2020}
Proen\, {c}a, H.M., van Leeuwen, M.: Interpretable multiclass classification by
  {MDL}-based rule lists. Information Sciences  512,  1372--1393 (Feb 2020),
  \url{https://www.sciencedirect.com/science/article/pii/S0020025519310138}

\bibitem{Rissanen1978}
Rissanen, J.: Modeling by shortest data description. Automatica  14(5),
  465--471 (1978)

\bibitem{sqs2012}
Tatti, N., Vreeken, J.: The long and the short of it: {Summarising} event
  sequences with serial episodes. In: Int. Conf. on Knowledge Discovery and
  Data Mining (KDD). pp. 462--470. ACM (2012)

\bibitem{KRIMP2011}
Vreeken, J., Van~Leeuwen, M., Siebes, A.: Krimp: mining itemsets that compress.
  Data Mining and Knowledge Discovery  23(1),  169--214 (2011)

\end{thebibliography}

\end{document}